\documentclass[9pt]{article}

\usepackage{amsfonts, amscd, mathrsfs, amsopn, bbm, bbold, multicol, relsize}
\usepackage{mathrsfs}
\usepackage[psamsfonts]{amssymb}
\usepackage{mathbbol}
\usepackage[mathscr]{eucal}
\usepackage[cmex10]{amsmath}
\usepackage[amsmath,thmmarks]{ntheorem}
\usepackage{mathtools}
\usepackage{cite}
\usepackage{tikz}
\usepackage{graphicx}

\theoremstyle{plain}
\theoremheaderfont{\itshape\bfseries}
\theorembodyfont{\normalfont\itshape}
\theoremseparator{:}
\theoremsymbol{}

\newtheorem{remark}{Remark}

\theoremstyle{nonumberplain}

\theoremstyle{nonumberplain}
\theoremheaderfont{\itshape}
\theorembodyfont{\normalfont}
\theoremseparator{:}
\theoremsymbol{}

\theoremstyle{plain}
\theoremheaderfont{\itshape\bfseries}
\theorembodyfont{\normalfont}
\theoremseparator{:}

\theoremstyle{nonumberplain}
\theoremsymbol{\rule{1ex}{1ex}}

\newlength\fheight
\newlength\fwidth
\def\S{ \mathcal{S} }               

\newcommand\restr[2]{{
  \left.\kern-\nulldelimiterspace 
  #1 
  \right|_{#2} 
  }}

\usepackage{spconf,amsmath,graphicx}
\usepackage[section]{placeins}
\usepackage{url}
\usepackage{relsize}
\usepackage{hyperref}
\usepackage{epstopdf}
\usepackage{enumerate}
\usepackage{algorithmicx}
\usepackage{algorithm}
\usepackage{amsmath}
\usepackage[noend]{algpseudocode}
\usepackage{float}
\usepackage{color}
\usepackage{makeidx}
\usepackage{bbm}
\usepackage{graphicx}
\usepackage{lipsum}
\usepackage{subcaption}
\usepackage[section]{placeins}

\usepackage{setspace}
\usepackage{lipsum}
\usepackage{adjustbox}
\usepackage{colortbl}
\usepackage{hhline}

\usepackage{multicol}
\definecolor{Gray}{gray}{0.88}
\definecolor{Grayy}{gray}{0.77}
\definecolor{Grayyy}{gray}{0.82}

\usepackage{lipsum}
\let\OLDthebibliography\thebibliography
\renewcommand\thebibliography[1]{
  \OLDthebibliography{#1}
  \setlength{\parskip}{0pt}
  \setlength{\itemsep}{0pt plus 0.3ex}
}

\title{LocUNet: Fast Urban Positioning Using
Radio Maps and Deep Learning}
\name{\c{C}a\u{g}kan Yapar$^{\dagger }$ \qquad Ron Levie$^{\ddagger}$ \qquad  Gitta Kutyniok$^{\ddagger\S}$ \qquad Giuseppe Caire$^{\dagger}$ 
\thanks{The long version of this paper is available at \cite{LocUNetArXiV}.}}

\address{\vspace{2mm} \large $^{\dagger}$ Institute of Telecommunication Systems, 
    TU Berlin, $^{\ddagger}$ Department of Mathematics, LMU München, \\
$^{\S}$Department of Physics and Technology, University of Troms{\o}
}
\newcounter{GittaCounter}

\newcounter{RonCounter}

\newcounter{CagkanCounter}

\newcounter{CagkanCounter2}

\newcounter{GCcounter}

\begin{document}
%
\maketitle
\begin{abstract}
This paper deals with the problem of localization in a cellular network in a dense urban scenario. Global Navigation Satellite Systems (GNSS) typically perform poorly in urban environments, where the likelihood of line-of-sight conditions is low, and thus alternative localization methods are required for good accuracy. We present LocUNet: A deep learning method for localization, based merely on Received Signal Strength (RSS) from  Base Stations (BSs), which does not require any increase in computation complexity at the user devices with respect to  the device standard operations, unlike methods that rely on time of arrival or angle of arrival information. In the proposed method, the user to be localized reports the RSS from BSs to a Central Processing Unit (CPU), which may be located in the cloud. Alternatively, the localization can be performed locally at the user. Using estimated pathloss radio maps of the BSs, LocUNet can localize  users with state-of-the-art accuracy and enjoys high robustness to inaccuracies in the radio maps. The proposed method does not require pre-sampling of the environment; and is suitable for real-time applications, thanks to the RadioUNet, a neural network-based radio map estimator. We also introduce two datasets that allow numerical comparisons of RSS and Time of Arrival (ToA) methods in realistic urban environments.
  
		  
\end{abstract}
\vspace{-2.2mm}
\vspace{-2.2mm}\vspace{-2.2mm}
\section{Introduction} \label{sec:Intro}
\vspace{-2.2mm}	
	
	The location information of a User Equipment (UE) is essential for many current and envisioned applications that range from emergency 911 services \cite{spect}, autonomous driving \cite{autonomousDriving}, intelligent transportation systems \cite{beyondGNSS}, proof of witness presence \cite{proofWitness}, 5G networks \cite{5G}, to name some.

	In urban environments, Global Navigation Satellite Systems (GNSS) alone may fail to provide a reliable localization estimate due to the lack of line-of-sight conditions between the UE and the GNSS satellites. In addition, the continuous reception and detection of GNSS signals is one of the dominating factors in battery consumption for hand-held devices. It is thus necessary to resort to other complementary means to  achieve the UE localization with the desired  high accuracy. 
	In cellular networks, the position of UE can be estimated by using different metrics that UE may report or the network can infer. The most prominent localization methods in the literature are based on Time of Arrival (ToA), Time Difference of Arrival (TDoA), Angle of Arrival (AoA) and Received Signal Strength (RSS) measurements.

\vspace{-2.2mm}\vspace{-2.2mm}\vspace{-2.2mm}
	\subsection{Received Signal Strength (RSS)}\label{subseq:RSS}
\vspace{-2.2mm}	\vspace{-0.4mm}
	
    Received signal strength quantifies the received power (averaged over a limited time interval of the beacon frames and over the signal bandwidth; hence it is not subject to small scale fluctuations, and coherent reception is not necessary) at a UE due to signals sent by a Base Station (BS). Since the transmit power of the BS on its beacon signal slots is fixed and known, the received signal strength is a function of the pathloss of the propagation between the BS and the UE. In fact, received signal strength measurements of beacon signals are routinely performed by UEs and reported to the system as ``received signal strength indicators'' (RSSI). Reporting RSS information is a standard feature in  most current wireless protocols. For example, this is used to trigger handovers and to enable UE-BS association for load balancing purposes. Therefore, exploiting RSS information for localization is attractive since it is a feature already built-in in the wireless protocols and does not require any further specific signal processing at the UE, whereas the time-based (ToA and TDoA) and angle-based (AoA) methods require high precision clocks and antenna arrays, respectively.

\vspace{-2.2mm}\vspace{-2.2mm}\vspace{-2.2mm}
	\subsection{Ranging-Based Methods}
	\vspace{-2.2mm}
	
	In ranging-based techniques, the distances between the UE and the BSs  are used to estimate the position of UE by lateration. Here, the distances are estimated  by using RSS or ToA measurements, based on a statistical signal attenuation or time delay model to estimate the distance between the UE and the BS. However, using such models is not appropriate in urban settings, since in practice the signal undergoes diverse propagation phenomena such as penetrations, reflections, diffractions, and wave guiding effects, due to the presence of obstacles in the environment. This leads to very large errors in the distance estimation (See Remark 2 in \cite{LocUNetArXiV} for a detailed discussion).  Several works, e.g. \cite{POCSgholami2011wireless,POCSHero,SDP,SDPR,ImpSDPR,BisecRob,Correntropy}, proposed methods to remedy this.  Nevertheless, these methods do not directly use a complete model of the propagation phenomena, and only partially alleviate the aforementioned problem. 

\vspace{-2.2mm}\vspace{-2.2mm}
	\subsection{Fingerprint-Based Methods}\label{subseq:FPMethods}
	\vspace{-2.2mm}

	Fingerprint-based methods, as opposed to ranging-based methods, do not impose any modeling assumption on the signal strength propagation. Instead, these methods rely on offline generated extensive databases of the measured signatures of the signal at different locations. Given a signal fingerprint, these methods infer the location of the UE by ``looking up'' a location with a similar fingerprint from the database (radio map, cf. Section \ref{sec:Preliminaries}) . Many fingerprint types, such as visual, radio wave (e.g. RSS or the baseband transfer function with complex channel coefficients, i.e., the Channel State Information (CSI)), or motion fingerprints, can be utilized for localization purpose \cite{FPSurvey}. In this paper we focus on RSS signatures, which are much more stable (cf. Subsection \ref{subseq:RSS}) and lower dimensional than CSI signatures. Fingerprint-based methods are well-known for outperforming the ranging-based techniques in complex urban environments \cite{laitinen2011comparison}.

\vspace{-2.2mm}\vspace{-2.2mm}
	\subsection{Physical Simulations of Radio Maps and RadioUNet}
	\vspace{-2.2mm}
	
	A major drawback of fingerprinting is the difficulty in generating and updating the radio maps. Fingerprinting requires a labor intensive, time consuming and expensive site surveying to generate the radio maps. Moreover, the reported positions of such measurements  suffer from the imprecision of GNSS in urban environments, the very problem the current paper is dealing with.

	A more feasible alternative for generating the radio map is using physical simulation methods such as ray-tracing \cite{RayTracingNew,WinPropFEKO}, thereby bypassing the extensive measurement campaigns. Based on an approximate model of the physical signal propagation phenomena, such simulations yield very accurate predictions, especially when accurate geometrical description of the urban environment is available. Many previous works proposed using simulated radio maps for localization, e.g. \cite{locRtEnt,deviceInd,MaherMalaney,propModel,peopleEffect,FPWinPropDL,wolfle2002enhanced,MLWinPropMultPath}.
	
	However, such simulations are computationally demanding and are thus not suitable for real time applications. Recently, an efficient deep learning-based method termed \emph{RadioUNet} \cite{RadioUNetTWC} (see Subsection \ref{RadioUNet}) was proposed by the authors of this paper. This algorithm enables the generation of high accuracy pathloss radio maps in much shorter time, and is thus adopted as an important building block of the current work.

\vspace{-2.2mm} \vspace{-2.2mm}
	\subsection{Machine Learning For Localization}
	\vspace{-2.2mm}
	
	Lately, several machine learning-based localization approaches have been proposed, e.g. \cite{LTE_DL,deepCReg,GarauBurguera2020},  see the recent surveys \cite{MLLocSurvey,wirelessDLSurvey}. To the best of our knowledge, none of the previous works benefited from the availability of accurate radio maps via physical simulations (or good approximations like RadioUNet), and relied solely on the RSS/CSI measurements of the signals from the BSs at the UE, or vice versa. We note that the radio maps serve straightforwardly as a means to assess a likelihood for the location of the UE, for each BS, e.g., by simply comparing the measured RSS with the radio map estimate at the location at question. As opposed to previous work, our presented method fully utilizes accurate radio map estimations instead of pathloss statistical models.

\vspace{-2.2mm}\vspace{-2.2mm}
\subsection{Our Contribution}	
\vspace{-2.2mm}
\textbf{1)} We propose an accurate and computationally efficient localization method, merely based on RSS measurements, which does not necessitate additional signal processing or hardware (e.g., calibrated antenna arrays) at the user devices. \textbf{2)} Using the recently developed RadioUNet, we can estimate radio maps very efficiently and accurately, by using the knowledge of the propagation environment, e.g., a map of the city, which we use in the presented work in order to achieve fast localization. \textbf{3)} The presented method relies on radio map (pathloss function) estimations and the RSS values from the device of interest. Based on these information and additional input features, the proposed neural network yields very accurate localization results and is robust to inaccuracies in radio map estimations. \textbf{4)} The proposed method allows for localization at the UE side, which can be used for, e.g. autonomous driving \cite{autonomousSelfLoc}. \textbf{5)} We introduce two synthetic datasets which are publicly available for the research community. 

\vspace{-2.2mm}\vspace{-2.2mm}
	\section{Preliminaries}\label{sec:Preliminaries}
\vspace{-2.2mm}
	\emph{Pathloss} (or \emph{large-scale fading coefficient}), quantifies the loss of wireless signal strength between a transmitter (Tx) and receiver (Rx) due to large scale effects. The signal strength attenuation can be caused by many factors, such as free-space propagation loss, penetration, reflection and diffraction losses by obstacles like buildings and cars in the environment. In dB scale\footnote{$(\cdot)_{\rm dB} := 10\log_{10}(\cdot)$.}, pathloss amounts to $\textup{P}_{\textup{L}} = (\textup{P}_{\textup{Rx}})_{\rm dB}-(\textup{P}_{\textup{Tx}})_{\rm dB}$,
	where $\textup{P}_{\textup{Tx}}$ and $\textup{P}_{\textup{Rx}}$ denote the transmitted and received locally averaged power (RSS) at the Tx and Rx locations, respectively. Notice that ``locally averaged'' power is defined as the energy per unit time averaged over time intervals of the order of a typical transmission slot in the underlying protocol (e.g., the duration of a Resource Block in 5G NR standard \cite{zaidi20185g}) and over the whole system bandwidth. Hence, the effect of the small-scale frequency selective fading is averaged out and only the frequency-flat pathloss matters.   
	
	A {\em radio map} $R(\mathbf{x}_1,\mathbf{x}_2)$, defines the pathloss in gray-level (pixel values between 0 and 1) between any two points $\mathbf{x}_1$ and $\mathbf{x}_2$ in the plane  (cf. \cite{LocUNetArXiV}). For fixed Tx position $\mathbf{x}_1 = \mathbf{x}_{\rm tx}$, the radio map is a function of the Rx position $\mathbf{x}_2 = \mathbf{x}_{\rm rx}$, i.e., it can be represented as a 2-dimensional image where the value of the pathloss $R(\mathbf{x}_{\rm tx},\mathbf{x}_{\rm rx})$ at each suitably discretized position $\mathbf{x}_{\rm rx}$ corresponds to a ``pixel'' of the image.


\vspace{-2.2mm}	\vspace{-2.2mm}
	\subsection{Radio Map Simulations}\label{subseq:RMapSims}
\vspace{-2.2mm}	
In this paper we considered two simulation models, namely \emph{Dominant Path Model} (DPM)\cite{DPM} and \emph{Intelligent Ray Tracing} \cite{IRT} with max. 2 interactions (IRT2), computed using the software WinProp \cite{WinPropFEKO} (cf. \cite{LocUNetArXiV} for the details).
\vspace{-2.2mm}\vspace{-2.2mm}
	\subsection{RadioUNet}
	\label{RadioUNet}
\vspace{-2.2mm}

	RadioUNet is a UNet \cite{UNet}-based pathloss radio map estimation method introduced in  \cite{RadioUNetConf,RadioUNetTWC}. In this paper we use the so called RadioUNet$_C$, which is a function that receives the map of the city and the location of a BS and returns an estimation of the corresponding radio map with a high accuracy, with root-mean-square-error of order of 1dB in various scenarios, and a run-time order of milliseconds on NVIDIA Quadro GP100 \cite{RadioUNetTWC}. RadioUNet is trained in supervised learning to match simulations of radio maps, using the RadioMapSeer dataset \cite{RadioMapSeer}. 
	
\vspace{-2.2mm}	\vspace{-2.2mm}
	\section{LocUNet}\label{sec:OurMethod}
\vspace{-2.2mm}

	Suppose that a user with location  $\mathbf{x}_U = (x_U, y_U)$ measures the strength of beacon signals (\emph{Non-interfering} identification signals), transmitted from a set of BSs $B_j$, $j=1,\ldots,J$, with known coordinates $\mathbf{x}_{B_j} = (x_{B_j},y_{B_j})$. Based on the relation between transmit/receive powers and the pathloss
	$(\textup{P}_{\textup{Rx}})_{\rm dB} = \textup{P}_{\textup{L}}+ (\textup{P}_{\textup{Tx}})_{\rm dB}$,
	the pathloss between the device and the BSs, $p_j$, $j=1,\ldots,J$, can be calculated, where we assume the small-scale fading effects are eliminated by averaging over time and system bandwidth. In our approach, the position of the UE is estimated based on: \textbf{1)}  The pathloss values $p_j$, $j=1,\ldots,J$, \textbf{2)} The estimations of the radio maps for each BS $R_j(x,y) := R(\mathbf{x}_{B_j}, (x,y))$, $j=1,\ldots,J$, computed via RadioUNet, \textbf{3)} The map of the urban environment, the locations of the BSs (which are fixed and known).  In order to input the above information \textbf{1)--3)}  to the UNet,
	it is first represented as a set of 2D images as follows: \textbf{1)} The RSS values $\{p_j\}$ are converted to gray-level as explained in \cite{RadioUNetTWC,LocUNetArXiV}.
	Each measured pathloss $p_j$  is represented as a 2D image $P_j(x,y)$ of the constant value $p_j$, i.e., for each $j$ this is an all-gray uniform image, but the level of gray differs for different indices $j$, \textbf{2)} 	Each radio map $R_j(x,y)$ is represented as a 2D image, with pixel value in gray-level. Radio maps are obtained by using RadioUNet, which takes the map of the urban environment, and the locations of BSs \cite{RadioUNetTWC}, \textbf{3)} The map of the urban environment is represented as a binary black and white image, where the interior of the buildings are white (pixel value $=1$), and the exterior is black (pixel value $=0$), \textbf{4)} The location of each BS $B_j$ is represented as a one-hot binary image, where the pixel at location $(x_{B_j},y_{B_j})$ is white, and the rest is black. These amount to $5+5+5+1=16$ input channels for $J=5$.
	
	The first part of LocUNet is a UNet variant \cite{UNet}, with average pooling, upsampling + bilinear interpolation, and Leaky ReLU as the activation function. We call the output feature map of the UNet, $H(x,y)$ a \emph{quasi-heat-map}, as its value at a point $(x,y)$ in the map quantifies the likelihood of the UE to be located at this point, while it can take negative values, due to LeakyReLu being the activation function of the network. The final layer of LocUNet computes the center of mass (CoM) $(\mu_x,\mu_y)$ of the quasi-heat-map $H(x,y)$, $\mu_x = \frac{\sum_{x=1}^{256}\sum_{y=1}^{256}x H(x,y)}{\sum_{x=1}^{256}\sum_{y=1}^{256} H(x,y)}$, $\mu_y = \frac{\sum_{x=1}^{256}\sum_{y=1}^{256}y H(x,y)}{\sum_{x=1}^{256}\sum_{y=1}^{256} H(x,y)}$,
	where 256 is the number of pixels along each axis. The architecture of LocUNet is summarized in Table \ref{table:LocUNet}.

	We measure the accuracy of the proposed method with mean absolute error \textbf{(MAE)}, which is the average 2D Euclidean distance between the estimated UE location and the ground-truth location. Namely, by
	$
	L(\mathbf{u},\tilde{\mathbf{u}}) = \frac{1}{|\mathcal{S}|} \sum_{k \in \mathcal{S}} ||\mathbf{u}^k - \tilde{\mathbf{u}^k}||$,
	where $\tilde{\mathbf{u}^k}:=(\mu_x^k,\mu_y^k)$ and $\mathbf{u}^k:=(x_U^k, y_U^k)$ denote the LocUNet estimation and the ground-truth of the $k$th instance of the dataset, and $\mathcal{S}$ is the entire training set. When using stochastic gradient descent for training, we take $\mathcal{S} = \mathcal{B}_m$, where $\mathcal{B}_m$ is the $m$th mini-batch  of the training set.

\begin{table}
\caption{\small Architecture of the first part of LocUNet (w/o final CoM layer). \emph{Resolution} is the number of pixels of the image in each feature channel along the $x,y$ axis. \emph{Filter size} is the number of pixels of each filter kernel along the $x,y$ axis.
		The input layer is concatenated in the last two layers before the CoM layer.} 
	\renewcommand{\arraystretch}{1}

	\centering
	\resizebox{\columnwidth}{!}{
		\begin{tabular}{c|c|c|c|c|c|c|c|c|c|c}
			\hline	
			\rowcolor{Grayyy} \multicolumn{11}{c}{\bfseries \quad \quad LocUNet}  \\
			\hline
			\rowcolor{Grayy} \bfseries Layer &\bfseries In &\bfseries 1 &\bfseries 2 &\bfseries 3 &\bfseries 4 &\bfseries 5 &\bfseries 6 &\bfseries 7 &\bfseries 8 &\bfseries 9\\
			\hline
			\cellcolor{Grayy} \bfseries Resolution & $256$ & $256$ & $128$ & $64$ & $64$ & $32$ & $32$ & $16$ & $16$ & $16$\\
			\hline
			\cellcolor{Grayy} \bfseries Channel & $16$ & $20$ & $50$ & $60$ & $70$ & $90$ & $100$ & $120$ & $120$ & $135$\\
			\hline
			\cellcolor{Grayy} \bfseries Filter size & $3$ & $5$ & $5$ & $5$ & $5$ & $5$ & $5$ & $3$ & $5$ & $5$\\
			\hline
			\rowcolor{Grayy} \bfseries Layer &\bfseries 10 &\bfseries 11 &\bfseries 12 &\bfseries 13 &\bfseries 14 &\bfseries 15 &\bfseries 16 &\bfseries 17 &\bfseries 18 &\bfseries 19\\
			\hline
			\cellcolor{Grayy} \bfseries Resolution & $8$ & $8$ & $4$ & $4$ & $2$ & $4$ & $4$ & $8$ & $8$ & $16$\\
			\hline
			\cellcolor{Grayy} \bfseries Channel & $150$ & $225$ & $300$ & $400$ & $500$ & $400+400$ & $300+300$ & $225+225$ & $150+150$ & $135+135$\\
			\hline
			\cellcolor{Grayy} \bfseries Filter size & $5$ & $5$ & $5$ & $5$ & $4$ & $5$ & $4$ & $5$ & $4$ & $5$\\
			\hline
			\rowcolor{Grayy} \bfseries Layer &\bfseries 20 &\bfseries 21 &\bfseries 22 &\bfseries 23 &\bfseries 24 &\bfseries 25 &\bfseries 26 &\bfseries 27 &\bfseries 28 &\bfseries 29\\
			\hline
			\cellcolor{Grayy} \bfseries Resolution & $16$ & $16$ & $32$ & $32$ & $64$ & $64$ & $128$ &   $256$ & $256$ & $256$\\
			\hline
			\cellcolor{Grayy} \bfseries Channel & $120+120$ & $120+120$ & $100+100$ & $90+90$ & $70+70$ & $60+60$ & $50+50$ & $20+20+16$ & $20+16$ & $1$\\
			\hline
			\cellcolor{Grayy} \bfseries Filter size & $3$ & $6$ & $5$ & $6$ & $5$ & $6$ & $6$ & $5$ & $5$ & -\\
			\hline

		\end{tabular}
	}
	\label{table:LocUNet}
		\vspace{-2.2mm}\vspace{-2.2mm}
\end{table}
	
\begin{remark}
The problem at hand looks like  classification of pixels as occupied by UE or not at first glance. Unfortunately, the number of classes in this formulation of the  problem is prohibitively large ($256^2$). Furthermore, the cross entropy loss used for classification tasks disregards the neighborhood of the pixels. Hence, opting for a classification approach for the localization problem presented in this paper is out of question. Another reasonable approach could be the straight forward regression approach, where instead of CoM layer, a fully connected layer would be used as the last layer. However, our experiments with this approach yielded a much inferior localization accuracy. Similar observations were made in \cite{EyeTracking}. 
\end{remark}

\vspace{-2.2mm}	\vspace{-2.2mm}\vspace{-2.7mm}
	\subsection{Training}
	\vspace{-2.2mm}
	
	We perform supervised learning on the training set. We use \emph{Adam} optimizer \cite{Adam} with an initial learning rate of $10^{-5}$ and decrease the learning rate by $10$ after $30$ epochs. We set the total number of epochs for training as $50$ and batch size as $15$. To avoid overfitting, we pick the network with the lowest validation error in the $50$ training epochs.
	 We used PyTorch for implementation\footnote{The code is available at \url{https://GitHub.com/CagkanYapar/LocUNet}. For reproducibility, see the compute capsule at \url{https://codeocean.com/capsule/7149386/tree}}.

	 \vspace{-5.2mm}
	\subsection{Datasets}\label{subseq:Dataset}
\vspace{-2.7mm}

We introduce the \emph{RadioLocSeer Dataset} of 99 city maps, $J=5$ BS locations for each, and corresponding simulated radio maps (By WinProp \cite{WinPropFEKO} and RadioUNet); and  \emph{RadioToASeer Dataset}, which provides ToA information based on the dominant path model (cf. \cite{DPM,LocUNetArXiV}) for the same maps of \emph{RadioLocSeer Dataset}, to allow for comparison between RSS and ToA ranging methods in urban scenario. Evaluating the ToA ranging-based methods on this dataset yields quasi-upper bounds for their performances. For details on the datasets \footnote{The datasets are available at \url{https://RadioMapSeer.GitHub.io/LocUNet.html}.} see \cite{LocUNetArXiV} .

   \begin{remark}
   To the best of our knowledge, there exists no publicly-available dataset of measured radio maps that represent the signal strength or ToA with a fine resolution, which is crucial for the localization task. As mentioned before, establishing real radio maps requires very expensive site surveying efforts, which is beyond the means of academic institutions. We note that many other previous works in communications relied on synthetically generated ground-truths by ray-tracing, e.g. \cite{locRtEnt,deviceInd,MaherMalaney,propModel,peopleEffect,FPWinPropDL,wolfle2002enhanced,MLWinPropMultPath,blockage, beamVehic, remoteChInference, mmBeamSel}.  \vspace{-2.2mm}
   \end{remark}

\vspace{-2.2mm}\vspace{-2.2mm}
	\section{Numerical Results}\label{seq:Numerical}
\vspace{-2.2mm}\vspace{-2.2mm}

	In this Section, we demonstrate the performance of LocUNet by numerical simulations. We assess the accuracy of LocUNet and the compared methods on the test set, namely, by MAE  with $\mathcal{S} = \mathcal{T}$, where $\mathcal{T}$ is the test set. LocUNet takes predicted radio maps, Tx location maps, pathloss measurements, and city map as input features. All of the compared algorithms with CPU implementation were run on an Intel Core i7-8750H, and LocUNet was run on a GPU of NVIDIA Quadro RTX 6000.

\vspace{-2.2mm}\vspace{-2.2mm}
	\subsection{LocUNet Scenarios}\label{subseq:Scenarios}
\vspace{-2.2mm}	
	
	We present three different scenarios to showcase the performance of LocUNet (and the compared algorithms) under different degrees of accuracy of the used radio map with respect to the true radio map, from which the pathloss measurements are reported. \textbf{DPM:}In this very optimistic scenario, we assume the availability of the ground truth (DPM simulations) radio maps to train RadioUNet. Hence, LocUNet enjoys having access to very high accuracy radio maps, where the inaccuracy of the available radio maps with respect to true radio maps is solely due to the prediction error of RadioUNet. \textbf{DPMToIRT2:}In this scenario, RadioUNet is trained on DPM simulations to generate fast DPM radio map predictions with good accuracy, while the pathloss measurements stem from IRT2 simulations, which are quite different from simulations/RadioUNet predictions of the dominant path model, of the same environment map. \textbf{DPMToIRT2Cars:} As in the previous scenario, RadioUNet is trained on DPM simulations, and the true pathloss measurements are taken from IRT2 simulations. Here, the IRT2 simulations are run on the same city map, but with  additional obstructions, cars, which were not present in the environment map (cf. \cite{LocUNetArXiV} for details), that RadioUNet used to estimate the radio maps of BSs. In this setting, the quality of the radio maps available to LocUNet is of the lowest accuracy.

\begin{remark}
The following numerical results are obtained with the above mentioned high computational power, which cannot be available at a usual UE. Hence, they showcase the localization performance of the LocUNet when run at a central unit (e.g. in the cloud), based on the reported RSS values of the UE for each BS. We hope that adapting the RadioUNet and the proposed LocUNet to the hardware limitations of a UE (so that the proposed method can be performed locally), e.g. by \emph{knowledge distillation} \cite{knowledgeDist}, won't degrade the overall performance remarkably. The investigation of this problem is however outside the scope of the current paper and is left as future work.   
\end{remark}	

\vspace{-2.2mm}\vspace{-2.2mm}
	\subsection{Comparison with Pathloss Fingerprint-Based Methods}
\vspace{-2.2mm}	
	
	In Table \ref{table:MAERunTimesFP}, we compare LocUNet with the competing fingerprint-based methods under the scenarios described in Subsection \ref{subseq:Scenarios}.
	We compare with two fingerprint-based methods, namely, \emph{k-nearest neighbors (kNN)} method \cite{RADAR} and an adaptive kNN variant \cite{adapKNN}. We determined the $k$ values of the kNN algorithm by coarse grid-search. We observe that LocUNet provides the best accuracy among the fingerprint-based methods for all the scenarios, and LocUNet is especially good at dealing with inaccuracies in the radio map estimations, as witnessed in the results of the DPMToIRT2Cars scenario. 

\vspace{-2.2mm}	\vspace{-2.2mm}
	\subsection{Comparison with ToA Ranging-Based Methods}\label{subseq:NumericalToA}
\vspace{-2.2mm}	

	We also compare our method with the state-of-the-art ToA ranging-based algorithms, which assume that it is not possible to distinguish whether a link between the UE and BS is in line-of-sight (LOS) or non-line-of-sight (NLOS). The difference between ranging with ToA measurement of the ray (range is calculated as: ToA $\times$ the speed of light) and the true geographical distance between UE and BS is called the \emph{NLOS bias}. See \cite{LocUNetArXiV} for the details on the choice of the parameters of the compared methods. In Table \ref{table:MAERunTimesToA} we show the numerical results of the methods based on ToA ranging. We also show the performance of the LocUNet scenario DPM. Note that both LocUNet and the ToA methods are implemented in the DPM setting, but the ToA methods use an additional information about ToA.  We observe that in this ideal scenario for LocUNet, MAE is $4.73$m, where the best result among ToA ranging-based methods is $7.16$m, which serves as a quasi-lower bound for ToA-based methods, due to the extremely optimistic \emph{RadioToASeer Dataset} \cite{LocUNetArXiV}.

	\begin{table}[!t]
	\caption{\small Comparison with fingerprint-based methods.} \label{table:MAERunTimesFP}
	\vspace{-2.2mm}	
		\renewcommand{\arraystretch}{1}

		\centering
		\scalebox{0.72}{
		\begin{tabular}{ccc}
			\hline	
			\rowcolor{Grayy}  {\bfseries  Algorithm }& \bfseries MAE & \bfseries Run-Time (ms)\\
			
			\rowcolor{Grayyy} \multicolumn{3}{c}{\bfseries \quad \quad DPM }  \\
			\hline
			kNN \cite{RADAR} (k=16)  & $7.01$& $\sim 20$\\
			\hline
			Adaptive kNN \cite{adapKNN} (avg. k=$2.50$)  & $7.49$& $\sim 20$\\
			
			\hline
			Proposed LocUNet &  $\mathbf{4.73}$& $\mathbf{\sim 5}$\\
			\hline
			\rowcolor{Grayyy} \multicolumn{3}{c}{\bfseries \quad \quad DPMToIRT2}  \\
			\hline
			kNN \cite{RADAR} (k=250)  & $23.38$& $\sim 20$\\
			\hline
			Adaptive kNN \cite{adapKNN} (avg. k=$6.55$)  & $25.39$& $\sim 20$\\
			\hline

			Proposed LocUNet &  $\mathbf{\mathbf{9.48}}$& $\mathbf{\sim 5}$\\
			\hline

			\hline
			
			\rowcolor{Grayyy} \multicolumn{3}{c}{\bfseries \quad \quad DPMToIRT2Cars}  \\
			\hline
			kNN \cite{RADAR} (k=300)  & $27.19$& $\sim 20$\\
			\hline
			Adaptive kNN \cite{adapKNN} (avg. k=$8.51$)  & $29.51$& $\sim 20$\\
			\hline
			
			Proposed LocUNet  &  $\mathbf{13.15}$& $\mathbf{\sim 5}$\\

			\hline
		\end{tabular}
		}
		
		\vspace{-1.7mm}
	\end{table}

	\begin{table}[!t]
		\caption{\small Comparison with ToA ranging-based methods.} \label{table:MAERunTimesToA}
		\vspace{-2.2mm}	
	\renewcommand{\arraystretch}{1}

	\centering
	\scalebox{0.68}{
	\begin{tabular}{ccc}
		\hline	
		\rowcolor{Grayy}  {\bfseries  Algorithm }& \bfseries MAE & \bfseries Run-Time (ms)\\
		\hline
		
		\hline		
		POCS \cite{POCSgholami2011wireless,POCSHero} & $37.89$& $\mathbf{\sim 15}$\\
		\hline		
		SDP  \cite{SDP} & $\mathbf{7.16}$& $\sim 600$\\
	
		\hline
		Robust SDP 1  \cite{SDPR} & $7.55$& $\sim 600$\\
	
		\hline
		Robust SDP 2 \cite{ImpSDPR} & $7.63$& $\sim 600$\\
	
		\hline
		Bisection-based robust method \cite{BisecRob} & $9.49$& $\sim 16$\\
		\hline
		Max. correntropy criterion method \cite{Correntropy} & $12.45$& $\sim 30$\\
		\hline\hline
		Proposed LocUNet DPM &  $\mathbf{4.73}$& $\mathbf{\sim 5}$\\
		
	\end{tabular}
	}

\end{table}

\vspace{-2.2mm}	
\newpage

\bibliographystyle{IEEEbib}

{\scriptsize
\bibliography{pub}
}

\end{document}